\documentclass[11pt]{article}

\PassOptionsToPackage{table}{xcolor}
\usepackage[final]{acl}
\usepackage[normalem]{ulem}

\usepackage[T1]{fontenc}
\usepackage{times}
\usepackage{latexsym}
\usepackage{graphicx}
\usepackage{booktabs}
\usepackage{multirow}
\usepackage{colortbl}
\usepackage{makecell}
\usepackage{tabularx}
\usepackage{amsmath}
\usepackage{enumitem}
\usepackage{float}
\usepackage{dblfloatfix}
\usepackage{url}
\usepackage{microtype}
\usepackage{placeins}

\definecolor{darkblue}{HTML}{1F4788}
\definecolor{coral}{HTML}{FFD4D4}
\definecolor{lightcyan}{HTML}{D4F1F4}
\definecolor{lightyellow}{HTML}{FFF9E6}
\definecolor{headerblue}{RGB}{220,230,245}
\definecolor{rowgray}{RGB}{247,247,247}

\title{IRTIQA: Evolving in English, Evaluated in Urdu — Does Agent Self-Evolution Transfer from English to Urdu and Code-Switched Input?}

\author{
Abdullah Hashmat \\
\texttt{abdullah.hashmat02@gmail.com}
\And
Danish Javed$^{1}$ \\
\texttt{28100421@lums.edu.pk}
\AND
Agha Ali Raza$^{1}$ \\
\texttt{agha.ali.raza@lums.edu.pk}
\And
Usman Naseem$^{2}$ \\
\texttt{usman.naseem@mq.edu.au}
\AND
{\normalfont $^{1}$Lahore University of Management Sciences, Pakistan} \\
{\normalfont $^{2}$Macquarie University, Australia}
}

\begin{document}
\maketitle
\begin{abstract}

As LLM agents increasingly learn from their own experience, the tasks they learn from and are evaluated on are almost always in English. We ask whether self-evolution transfers from English to Urdu, and release IRTIQA, a benchmark of aligned English, Urdu-translated, and Urdu code-switched variants of DROP, SNIPS, AppWorld, and SkillsBench, built with a matrix-language code-switching pipeline and human verification. Using IRTIQA, we evaluate three self-evolving frameworks, Tool-R0 (self-play RL), AgentEvolver (experience-guided RL), and Autogenesis (protocol-level evolution), against their unevolved backbones. Self-evolution often raises capability: AgentEvolver lifts AppWorld goal completion from near zero to 20--36\% across all conditions, and Autogenesis improves SkillsBench pass rate from 16.48\% to 22.22\%. But it rarely closes the language gap. Evolved systems remain stronger in English, by up to 0.26 DROP F1 and 30 AppWorld goal points, the gap is largest on multi-step tasks, and code-switching behaves as a distinct condition rather than degraded translation. On a shared Gemma-4 31B backbone, the gap exists before evolution, and RL-based evolution can hurt Urdu reasoning more than English. Language parity must therefore be an explicit design target, as it does not emerge from self-improvement alone. Data and code are available at \href{https://github.com/Hashmat02/IRTIQA}{here}.
\end{abstract}

\section{Introduction}

Self-evolving agents reduce reliance on human-annotated trajectories by
generating their own tasks, solutions, and feedback, and using the results back into their subsequent behavior. Frameworks like Tool-R0 \cite{acikgoz2026tool} use self-play between a task generator and a solver, while AgentEvolver \cite{zhai2025agentevolver} combines task generation, experience reuse, and fine grained credit assignment, EvoSkills \cite{zhang2026coevoskills} co-evolves a skill generator model and its surrogate verifier to excel at a certain task. ASI-Evolve \cite{xu2026asi} frames evolution as a loop of learning, experimentation, and analysis, more along the lines of an AI scientist. What these systems have in common is that the model under training both produces its own tasks and solutions and also decides whether they succeeded, through self-generated rewards, verifiers and critiques. Hence, no external human signal enters the improvement loop.

This closed loop makes language more than a surface level shift, where a model that reasons poorly in Urdu will also generate weaker Urdu tasks and judge Urdu solutions less reliably, so self-evolution may widen the language gap instead of closing it. In a supervised pipeline, a translation error corrupts one example. In a self-evolving loop, the model that misreads an input is often the model that verifies the resulting trajectory, so a language-induced error could in principle be accepted, stored as experience or skill, and reinforced in later rounds \cite{kamoi2024can}. Yet every framework above is evolved and evaluated almost entirely in English. This leads to a foundational question, when an agent has taught itself in English, how much of what it learned survives when the input arrives in another language? We study this transfer question directly and leave measuring propagation during evolution to future work. This matters in deployment, where many users interact not in monolingual English but through code-switching \cite{dogruoz2021survey} as well. Moreover, multilingual model competence does not necessarily guarantee reliable representations under mixed-language input \cite{winata2021multilingual}. Existing multilingual robustness studies offer limited guidance, since they evaluate single-pass models rather than agents whose abilities were acquired through an iterative generate--verify--refine process.

In our study, we answer this question for English$\rightarrow$Urdu. Urdu is a widely spoken, low-resource and non-Latin script language with pervasive English code-switching. We introduce \textbf{IRTIQA}\footnote{IRTIQA, Urdu for `evolution' or `gradual ascent,' named for the self-evolution process.}, a benchmark in which every instance appears in three aligned conditions sharing an aligned gold answer: the original English, a human-audited Urdu translation, and a linguistically constrained Urdu code-switched variant in which Urdu supplies the grammatical frame and English supplies content words. Holding the task and answer fixed across conditions attributes any performance change to the language transformation alone. On IRTIQA we evaluate three English evolved frameworks spanning distinct evolution mechanisms: Tool-R0, AgentEvolver, and Autogenesis, against their unevolved base models. The evaluation datasets cover  structured language understanding (SNIPS \cite{coucke2018snips}), numerically grounded reading comprehension (DROP \cite{dua2019drop}), multi-step skills execution (SkillsBench \cite{li2026skillsbench} and AppWorld~\cite{trivedi2024appworld}) for everyday multi-app agent tasks. The paired design separates what self-evolution adds from what the backbone model already had, in each language. To observe backbone model effect, we run all frameworks on a shared Gemma-4 31B model, this controls for differences between their native backbones. In both settings, the English--Urdu gap is already present in the unevolved backbone, and no framework closes it. Our contributions are:

\begin{itemize}[
    leftmargin=*,
    itemsep=1pt,
    topsep=2pt,
    parsep=0pt,
    partopsep=0pt
]
    \item \textbf{IRTIQA}, a released benchmark of aligned English, Urdu,
    and Urdu code-switched variants of four agent benchmarks. It is produced by a reusable translation and code-switching pipeline with human quality control.

    \item A matrix language constrained code-switching method that preserves Urdu morphology, function words, and word order while substituting English content words, rather than performing lexical substitution.

    \item The first cross-framework study of whether agent self-evolution
    transfers from English to Urdu. It quantifies degradation by
    transformation type and task family against paired unevolved baselines, under both native and shared backbones.

    \item An outcome-level failure analysis by answer type, domain, intent, and task difficulty, with paired significance tests. It shows where language gaps persist and how they vary with task difficulty.
\end{itemize}

\section{Related Work}

\subsection{Self-evolving agents}

Self-evolving agents generate their own curricula, solutions, and feedback.
Besides the frameworks we evaluate in Section~\ref{sec:frameworks},
Tool-R0 \cite{acikgoz2026tool}, AgentEvolver \cite{zhai2025agentevolver},
and Autogenesis \cite{zhang2026autogenesis}, adjacent systems extend the same template in other directions: EvoSkills \cite{zhang2026coevoskills} co-evolves skill generators and surrogate verifiers, ASI-Evolve \cite{xu2026asi} evolves data, architectures, and learning algorithms, and InfiAgent \cite{yu2025infiagent} restructures hierarchical multi-agent systems through dynamic DAGs. All of these systems are evolved and evaluated on English benchmarks, and to our knowledge none is tested on translated or code-switched input,
which is the gap IRTIQA addresses.

\subsection{Reliability of self-generated feedback}
LLMs often fail to evaluate their own outputs reliably, revising correct
answers or reinforcing flawed reasoning
\cite{huang2024large,kamoi2024can,tyen2024llms}, and reported
self-correction gains frequently depend on external feedback, stronger
verifiers, or explicit verification signals rather than genuine
self-verification \cite{wu2024large}. Self-refinement can also exhibit
self-preference bias toward a model's own generations \cite{xu2024pride,wataoka2024self}. In self-evolving agents these failures
compound: an unreliable judgment does not merely mis-score one output but
admits it into the experience buffer or skill library, which is why
EvoSkills, AgentEvolver, and ASI-Evolve all build in filtering or
verifier-based mechanisms. All of this evidence, however, comes from
English. We ask the transfer question this literature leaves open, whether capabilities produced by English evolution survive translation and code-switching, and leave contamination during evolution itself to future work.

\subsection{Multilingual evaluation of agentic systems}
Agent benchmarks remain predominantly English-centric. ToolBench
\cite{qin2024toolllm}, API-Bank \cite{li2023api}, NexusRaven
\cite{srinivasan2023nexusraven}, BFCL \cite{patil2025berkeley}, SkillsBench
\cite{li2026skillsbench}, and AppWorld \cite{trivedi2024appworld} evaluate
tool use, function calling, skill execution, and long-horizon interaction
through English instructions and schemas, and the reasoning benchmarks
self-evolving frameworks report on, DROP \cite{dua2019drop}, MMLU
\cite{hendrycks2021measuring}, GAIA \cite{mialon2024gaia}, and GPQA
\cite{rein2023gpqa}, are likewise English. On the other side,
code-switching benchmarks such as GLUECoS \cite{khanuja2020gluecos} and
LinCE \cite{aguilar2020lince} target static NLU tasks with single-pass
classification and sequence labeling. They involve no tools or environment state, and the model does not improve itself. The two literatures therefore do not
meet, and to our knowledge no existing benchmark evaluates an agent whose competence was
acquired through self-evolution on input in another language or register.
IRTIQA sits at that intersection, pairing every evolved
system with its own backbone and with a shared backbone so that framework effects are separable from
base-model multilingual competence.

\begin{figure*}[t]
    \centering
    \includegraphics[width=.95\textwidth]{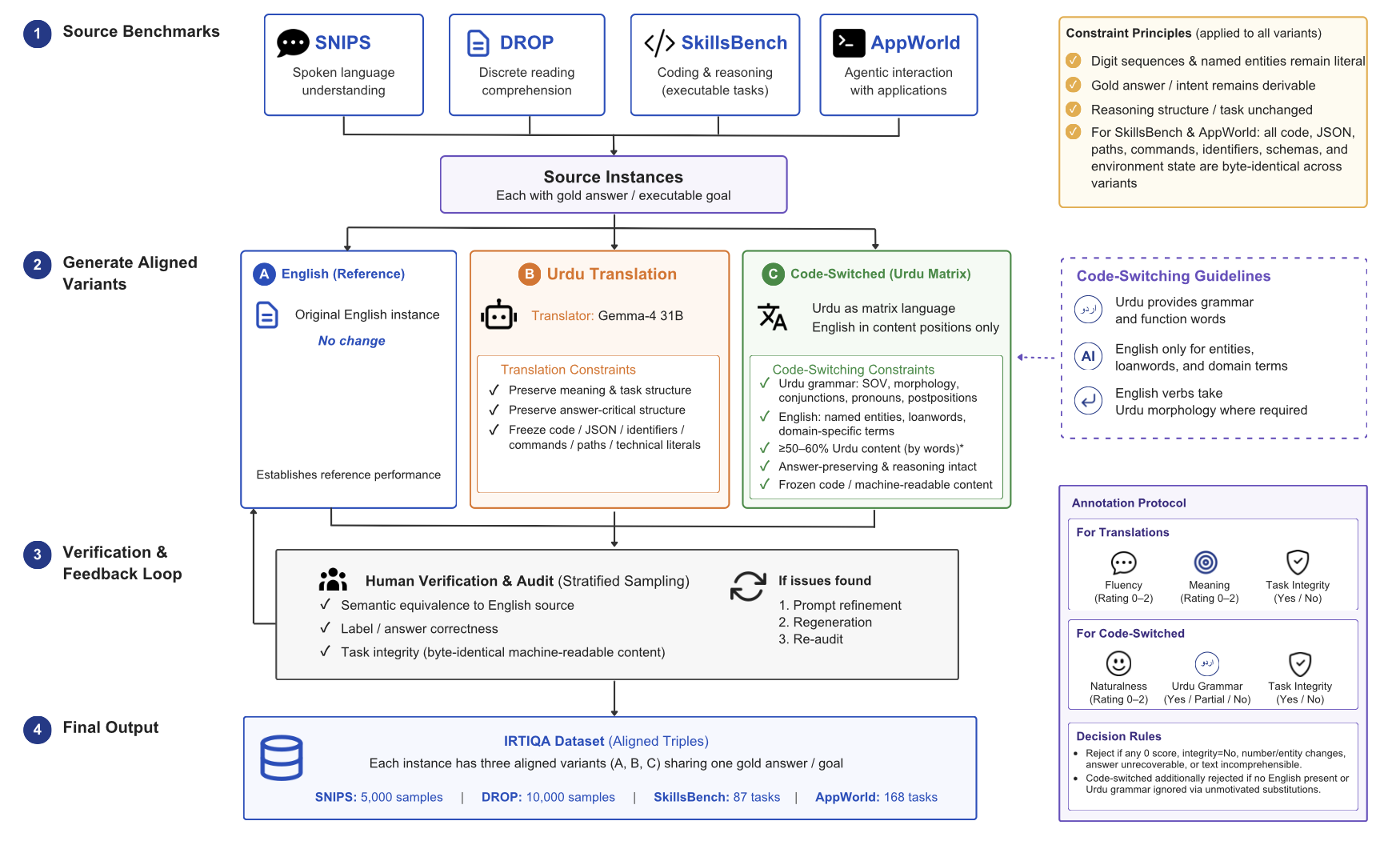}
    \caption{Overview of the methodology and evaluation pipeline.}
    \label{fig:methodology-overview}
\end{figure*}

\section{Methodology}

Figure~\ref{fig:methodology-overview} shows an overview of our methodology and evaluation pipeline, including dataset transformation, human auditing, framework evaluation, and cross-lingual analysis. We run Framework evaluation once in each framework's released configuration and on a shared Gemma-4 31B backbone (Section~\ref{sec:shared-backbone-setup}).

\subsection{Evaluation Dataset Variants}
\vspace{-4pt}
For every source instance in SNIPS, DROP, SkillsBench, and AppWorld, we construct aligned \textbf{English}, \textbf{Urdu-translated}, and \textbf{Urdu code-switched} versions sharing the same gold answer (with an aligned Urdu script form for DROP) or executable goal. For executable tasks, code blocks, paths, JSON keys, commands, tool schemas, identifiers, and environment state remain unchanged. The design therefore changes the user facing language while preserving the underlying task.

\paragraph{Translation.}
On a held-out subset, we compared instruction tuned multilingual LLMs (GPT-4o-mini\footnote{\url{https://developers.openai.com/api/docs/models/gpt-4o-mini}}, Gemma-4 31B\footnote{\url{https://huggingface.co/google/gemma-4-31B}}, and LLaMA-3 70B\footnote{\url{https://huggingface.co/meta-llama/Meta-Llama-3-70B}}) alongside dedicated matchine translation (MT) systems (SeamlessM4T, NLLB, and Google Translate). Human annotators scored fluency, adequacy, and cultural and linguistic correctness. Gemma-4 31B ranked highest and was chosen to generate the final set, reducing translation quality as a confound. General prompts translate natural language fields while preserving answer critical structure, while the SkillsBench template additionally freezes code, paths, JSON, commands, identifiers, and technical literals (Appendix~\ref{app:translation}).

\paragraph{Code-switching.}
A language specific prompt fixes Urdu as the grammatical matrix and restricts English to content positions under the constraints in Section~\ref{sec:cs}. Appendix~\ref{app:codeswitch} provides the full template and rules used for code-switching.

\paragraph{Data verification and finalization.}
Annotators audit a stratified sample for semantic equivalence, label correctness, and task integrity (Section~\ref{sec:annotation}, agreement is mentioned in Appendix Table~\ref{tab:iaa}). We identify recurring errors and refine the prompts and constraints, after which the data is regenerated. 

\subsection{Linguistically Grounded Code-Switching}
\vspace{-4pt}
\label{sec:cs}
Unlike random lexical substitution, our generation follows explicit linguistic constraints. Urdu supplies SOV word order, verb morphology, conjunctions, pronouns, and postpositions and remains the majority language, at least 50--60\% of content vocabulary in the general template. English is limited to named entities, established loanwords, and domain specific terms. English verbs receive Urdu morphology where needed. Where as digit sequences and English named entities remain literal, the answer or intent remains derivable, and the reasoning structure is preserved. Generation and human audit apply the same constraints (Appendix~\ref{app:codeswitch}).

\subsection{Annotation Protocol}
\vspace{-4pt}
\label{sec:annotation}
Annotators independently compare each sampled variant with its English source and original answer or label. Translations are evaluated where fluency and meaning preservation are give a score between ($0$--$2$). Task integrity is scored in a binary manner (\textsc{Yes}/\textsc{No}), where as code-switched items are judged for naturalness ($0$--$2$), and Urdu grammar fidelity using (\textsc{Yes}/\textsc{Partial}/\textsc{No}) with the same integrity criterion. A sample is rejected if any score is zero, integrity fails, a number or entity changes, or the text is incomprehensible. Otherwise, translations scoring 1 on both fluency and meaning are flagged, code-switched items are flagged when naturalness is 1 and grammar fidelity is \textsc{Partial}. Code-switched outputs are also rejected if they contain no English or ignore Urdu grammar through unmotivated substitutions. Appendix~\ref{app:annotation-rubrics} gives the complete rubrics and decision tables. Inter-annotator agreement, measured with Gwet's AC1~\cite{gwet2008computing}, is given in Appendix Table~\ref{tab:iaa}. Final decisions follow a majority vote, and ties are resolved toward rejection.

\section{Experimental Setup}

We compare each evolved system with its un-evolved backbone under identical interfaces and language conditions. Importantly, multilingual inputs are introduced only at evaluation time. Both the released systems and our shared backbone reruns are evolved in English. English to Urdu differences measure English--Urdu degradation, translation to code-switching differences isolate transformation type, and within-pair changes estimate the gain associated with self-evolution. 

\subsection{Datasets}
\vspace{-4pt}
\label{sec:datasets}

\begin{figure*}[t]
    \centering
    \includegraphics[width=1.0\textwidth]{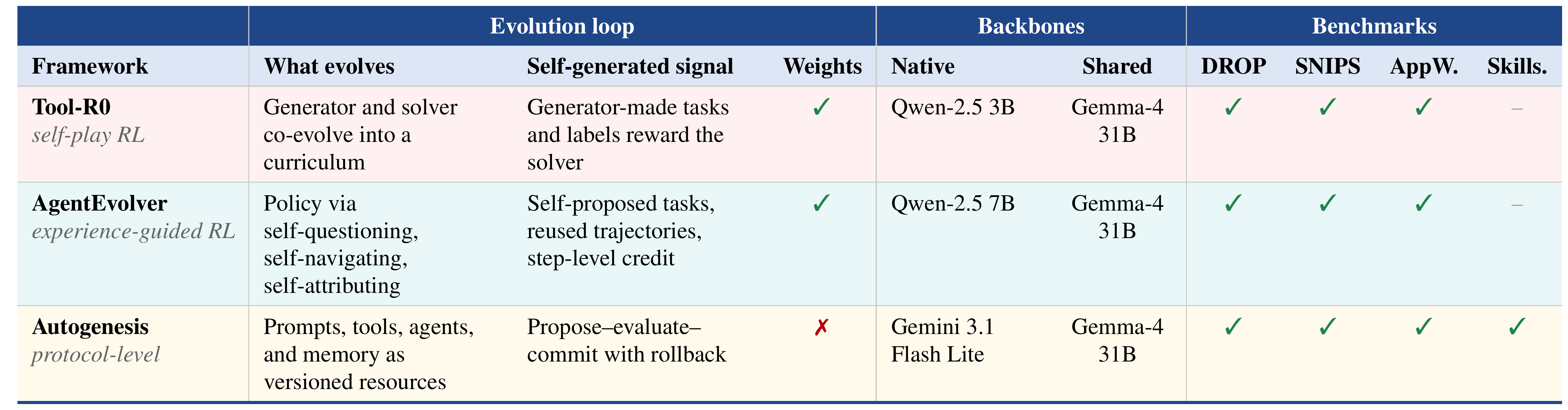}
    \caption{Evaluated self-evolving frameworks and autonomous improvement mechanisms.}
    \label{fig:frameworks}
\end{figure*}









We use four complementary benchmarks, each in all three aligned variants. \textbf{SNIPS}\footnote{\url{https://huggingface.co/datasets/DeepPavlov/snips}} evaluates intent detection and slot filling, we evaluate 5,000 instances per condition and report intent accuracy averaged over evaluation trials. A fixed 700 instance paired subset used for inferential analysis is described in Appendix~\ref{app:statistics}. \textbf{DROP}\footnote{\url{https://huggingface.co/datasets/ucinlp/drop}} requires discrete reasoning over passages, we evaluate 10,000 instances per condition using Exact Match (EM) and token-level F1. While \textbf{SkillsBench}\footnote{\url{https://huggingface.co/datasets/benchflow/skillsbench}} evaluates tool-augmented execution across all 87 tasks in roughly 20 professional domains, its deterministic verifier yields trial-averaged pass rates by Core, Extended, and Extreme difficulty. We compute the overall pass rate over all tasks and stratified rates within each tier.\textbf{AppWorld}\footnote{\url{https://appworld.dev/appworld/}} measures interactive completion in simulated everyday applications, we evaluate all 168 tasks of the split under greedy decoding and preserve application state, entities, constraints, and tools across conditions. Normalized reward measures partial progress and goal completion measures end-to-end success. 

\subsection{Evaluation Metrics}
\vspace{-4pt}
\label{sec:metrics}

We follow each benchmark's standard protocol. Each DROP record stores the original gold answer together with an aligned Urdu-script version (e.g., \texttt{284} is also stored in Eastern Arabic digits), so answers given in Urdu script are not penalized for script. We treat F1 as the primary DROP metric because it credits partial overlap, and report EM alongside it. Table~\ref{tab:evaluation-metrics} in the appendix summarizes the metrics.

\subsection{Self-Evolving Frameworks}
\label{sec:frameworks}
\vspace{-4pt}
We evaluate self-play RL, experience-driven RL, and protocol-level evolution (Figure~\ref{fig:frameworks}). 

\textbf{Tool-R0} \cite{acikgoz2026tool} trains without any external data. A task generator and a tool-using solver play against each other, and the generator's tasks get harder but stay solvable, forming an automatic curriculum. We pair the released Qwen-2.5 3B system with the corresponding base model. \textbf{AgentEvolver} \cite{zhai2025agentevolver} combines \emph{self-questioning}, \emph{self-navigating}, and \emph{self-attributing} to generate tasks, reuse experience, and assign trajectory level credit, here we use its Qwen-2.5 7B system with its base model. \textbf{Autogenesis} \cite{zhang2026autogenesis} evolves versioned prompts, agents, tools, environments, and memory through propose--evaluate--commit updates with rollback. We compare its Gemini 3.1 Flash Lite instantiation with the same backbone without Autogenesis. Official implementations, checkpoints, interfaces, and evaluation procedures are retained wherever supported.

\paragraph{Shared-backbone setting.}
\label{sec:shared-backbone-setup}
Evaluating frameworks on their native backbones confounds the evolution paradigm with model family and scale. We therefore re-instantiate all three frameworks on \texttt{google/gemma-4-31B-it} and compare each with its unevolved backbone. Evolution again relies solely on each framework’s English procedures. We keep the official hyperparameters for our runs. The two RL frameworks, Tool-R0 and AgentEvolver, update the Gemma weights using their English recipes. Autogenesis leaves the weights unchanged and evolves prompts, tools, and memory instead.
Every system, including the base, is decoded greedily with vLLM (Appendix Table~\ref{tab:gemma-settings}).
Note that Gemma-4 31B also generated the translated and code-switched variants, which may favor it on surface form in this setting.

\section{Results}
\subsection{Framework-Level Results}
\vspace{-4pt}
We report results by framework, comparing each evolved system with its own backbone under the same language condition. Section~\ref{sec:shared-backbone-results} shows this comparison on a shared backbone. While evolution is most effective when the backbone can already interpret the task, each framework shows a distinct English--Urdu pattern.

\subsubsection{Tool-R0: self-play evolution with Qwen-2.5 3B}
\vspace{-4pt}
Tool-R0 produces selective gains (Table~\ref{tab:toolr0-overall}). On DROP, F1 and EM rise in every condition, however, the evolved system still loses 0.263 F1 from English to translated Urdu, so self-play raises performance without closing the language gap. Aggregate SNIPS accuracy also improves, although several individual Urdu intents deteriorate (Section~\ref{sec:failure-analysis}). On AppWorld, reward changes are negligible and goal completion remains zero. Hence, we observe that Tool-R0 transfers to short-form reasoning and classification but does not acquire the planning capacity required for long-horizon application control.

\begin{table}[t]
\centering
\small
\setlength{\tabcolsep}{4pt}
\renewcommand{\arraystretch}{1.1}
\resizebox{\columnwidth}{!}{%
\begin{tabular}{lllrr}
\toprule
\textbf{Dataset} & \textbf{Metric} & \textbf{Variant}
& \textbf{Base} & \textbf{Tool-R0} \\
\midrule
\multirow{6}{*}{DROP}
& \multirow{3}{*}{F1}
& EN & 0.305 & 0.453 \\
& & UR-TR & 0.0997 & 0.190 \\
& & UR-CS & 0.1567 & 0.234 \\
\cmidrule(lr){2-5}
& \multirow{3}{*}{EM}
& EN & 0.264 & 0.394 \\
& & UR-TR & 0.071 & 0.136 \\
& & UR-CS & 0.125 & 0.190 \\
\midrule
\multirow{3}{*}{SNIPS}
& \multirow{3}{*}{Accuracy}
& EN & 90.71 & 91.70 \\
& & UR-TR & 70.57 & 72.90 \\
& & UR-CS & 82.14 & 83.30 \\
\midrule
\multirow{6}{*}{AppWorld}
& \multirow{3}{*}{Reward}
& EN & 0.02 & 0.03 \\
& & UR-TR & 0.00 & 0.00 \\
& & UR-CS & 0.00 & 0.01 \\
\cmidrule(lr){2-5}
& \multirow{3}{*}{Goal}
& EN & 0 & 0 \\
& & UR-TR & 0 & 0 \\
& & UR-CS & 0 & 0 \\
\bottomrule
\end{tabular}%
}
\caption{Tool-R0 compared with its Qwen-2.5 3B backbone
(\emph{Base}) across language conditions. SNIPS accuracy and
AppWorld goal completion are reported as percentages. DROP and SNIPS values are full-run trial averaged aggregates while AppWorld is a single greedy-decoded run on the 168 task split.}
\label{tab:toolr0-overall}
\end{table}

\subsubsection{AgentEvolver: experience-driven evolution with Qwen-2.5 7B}
\vspace{-4pt}
\begin{table}[t]
\centering
\scriptsize
\setlength{\tabcolsep}{3.5pt}
\renewcommand{\arraystretch}{1.08}
\resizebox{\columnwidth}{!}{%
\begin{tabular}{lllrr}
\toprule
\textbf{Dataset} & \textbf{Metric} & \textbf{Lang.}
& \textbf{Base} & \textbf{AgentEvolver} \\
\midrule

\multirow{6}{*}{DROP}
& \multirow{3}{*}{F1}
& EN    & 0.4345 & 0.534 \\
&       & UR-TR & 0.2337 & 0.340 \\
&       & UR-CS & 0.2271 & 0.313 \\
\cmidrule(lr){2-5}
& \multirow{3}{*}{EM}
& EN    & 0.361 & 0.462 \\
&       & UR-TR & 0.177 & 0.287 \\
&       & UR-CS & 0.169 & 0.271 \\
\midrule

\multirow{3}{*}{SNIPS}
& \multirow{3}{*}{Accuracy}
& EN    & 90.00 & 91.70 \\
&       & UR-TR & 84.71 & 87.80 \\
&       & UR-CS & 91.14 & 92.60 \\
\midrule

\multirow{6}{*}{AppWorld}
& \multirow{3}{*}{Reward}
& EN    & 0.06 & 0.37 \\
&       & UR-TR & 0.01 & 0.25 \\
&       & UR-CS & 0.02 & 0.30 \\
\cmidrule(lr){2-5}
& \multirow{3}{*}{Goal}
& EN    & 2.38 & 35.71 \\
&       & UR-TR & 0.00 & 20.24 \\
&       & UR-CS & 0.60 & 24.40 \\
\bottomrule
\end{tabular}%
}

\caption{AgentEvolver compared with its Qwen-2.5 7B backbone
(\emph{Base}) across English (EN), Urdu translation (UR-TR), and
Urdu code-switching (UR-CS). SNIPS accuracy and AppWorld goal
completion are reported as percentages.}
\label{tab:agentevolver-overall}
\end{table}

AgentEvolver provides the clear cross-task gains (Table~\ref{tab:agentevolver-overall}). On DROP, F1 improves by 0.100 in English, 0.106 under translation, and 0.086 under code-switching. SNIPS gains are largest in translated Urdu (+3.09 points), followed by English (+1.70) and code-switching (+1.46), showing positive transfer of experience reuse and trajectory-level credit assignment. The largest change occurs on AppWorld where goal completion rises from 2.38\% to 35.71\% in English, 0\% to 20.24\% under translation, and 0.6\% to 24.4\% under code-switching, with parallel reward gains (per-difficulty results in Appendix Table~\ref{tab:appworld-difficulty-full}). This turns a nearly non-functional backbone into an agent that completes a meaningful share of level-1 and level-2 tasks. Overall, English remains 11--15 points higher and no level-3 task is completed, so evolution lets the agent solve more tasks, but it still struggles with Urdu input and long multi-step tasks.

\subsubsection{Autogenesis: protocol-level evolution with Gemini}
\vspace{-4pt}
Autogenesis improves an already strong Gemini baseline on every aggregate (Table~\ref{tab:autogenesis-overall}). DROP gains are modest but consistent at +0.015, +0.015, and +0.021 F1 in English, translated Urdu, and code-switched Urdu. These gains are much smaller than the remaining English--translation gap (0.806 vs.\ 0.634 F1), so protocol refinement does not make reasoning language invariant.

Both systems approach ceiling on SNIPS with no significant cross-language differences. On AppWorld, goal completion rises by 7.73, 5.95, and 7.14 points and reward improves in every condition. SkillsBench improves consistently as well: the 3-trial overall pass rate rises from 16.48/12.64/14.18\% to 22.22/17.24/19.16\% in EN/UR-TR/UR-CS. Core gains exceed 11 points in every condition, while Extreme pass rate more than doubles throughout. Yet the evolved system retains a 4.98-point EN--UR-TR overall gap and solves fewer than 9\% of Extreme tasks in every language. Protocol-level evolution therefore helps even with a strong backbone without removing planning limits or the English--Urdu gap.

\begin{table*}[t]
\centering
\scriptsize
\setlength{\tabcolsep}{4pt}
\renewcommand{\arraystretch}{1.05}
\resizebox{\textwidth}{!}{%
\begin{tabular}{llrrrrrr}
\toprule
\textbf{Dataset} & \textbf{Metric}
& \multicolumn{2}{c}{\textbf{EN}}
& \multicolumn{2}{c}{\textbf{UR-TR}}
& \multicolumn{2}{c}{\textbf{UR-CS}} \\
\cmidrule(lr){3-4}
\cmidrule(lr){5-6}
\cmidrule(lr){7-8}
& & \textbf{Gemini base} & \textbf{Autogenesis}
  & \textbf{Gemini base} & \textbf{Autogenesis}
  & \textbf{Gemini base} & \textbf{Autogenesis} \\
\midrule

\multirow{2}{*}{DROP}
& F1
& 0.791 & \textbf{0.806}
& 0.619 & \textbf{0.634}
& 0.742 & \textbf{0.763} \\

& EM
& 0.712 & \textbf{0.727}
& 0.498 & \textbf{0.507}
& 0.651 & \textbf{0.673} \\

\midrule

SNIPS
& Accuracy (\%)
& 97.70 & \textbf{98.33}
& 98.00 & \textbf{98.67}
& 97.80 & \textbf{98.33} \\

\midrule

\multirow{2}{*}{AppWorld}
& Reward
& 0.710 & \textbf{0.783}
& 0.660 & \textbf{0.731}
& 0.640 & \textbf{0.706} \\

& Goal (\%)
& 64.29 & \textbf{72.02}
& 50.00 & \textbf{55.95}
& 52.98 & \textbf{60.12} \\

\bottomrule
\end{tabular}%
}
\caption{Autogenesis compared with the corresponding Gemini 3.1 Flash Lite baseline across English (EN), Urdu translation (UR-TR), and Urdu code-switching (UR-CS). Bold indicates the higher score.}
\label{tab:autogenesis-overall}
\end{table*}

\begin{table}[t]
\centering
\scriptsize
\setlength{\tabcolsep}{3pt}
\renewcommand{\arraystretch}{1.08}
\resizebox{\columnwidth}{!}{%
\begin{tabular}{llrrrr}
\toprule
\textbf{System} & \textbf{Variant} & \textbf{Core} & \textbf{Extended} & \textbf{Extreme} & \textbf{Overall} \\
\midrule
\multirow{3}{*}{Gemini}
& EN    & 55.56\% & 18.87\% & 3.57\% & 16.48\% \\
& UR-TR & 44.44\% & 15.09\% & 1.19\% & 12.64\% \\
& UR-CS & 50.00\% & 16.35\% & 2.38\% & 14.18\% \\
\midrule
\multirow{3}{*}{Autogenesis}
& EN    & \textbf{66.67\%} & \textbf{24.53\%} & \textbf{8.33\%} & \textbf{22.22\%} \\
& UR-TR & \textbf{55.56\%} & \textbf{19.50\%} & \textbf{4.76\%} & \textbf{17.24\%} \\
& UR-CS & \textbf{61.11\%} & \textbf{21.38\%} & \textbf{5.95\%} & \textbf{19.16\%} \\
\bottomrule
\end{tabular}%
}
\caption{SkillsBench verifier confirmed pass rate of Gemini 3.1 Flash Lite and Autogenesis by language variant and difficulty (Core/Extended/Extreme = easy/medium/hard). Bold marks Autogenesis improvements over the matched Gemini baseline. Values are averaged over 3 trials across all 87 tasks.}
\label{tab:skillsbench}
\end{table}
\subsection{Shared-Backbone Control: Gemma-4 31B}
\label{sec:shared-backbone-results}

\begin{table}[H]
\centering
\scriptsize
\setlength{\tabcolsep}{3.5pt}
\renewcommand{\arraystretch}{1.08}
\resizebox{\columnwidth}{!}{%
\begin{tabular}{llrrrrrr}
\toprule
& & \multicolumn{2}{c}{\textbf{DROP}} & \textbf{SNIPS} & \multicolumn{2}{c}{\textbf{AppWorld}} & \textbf{Skills.} \\
\cmidrule(lr){3-4}\cmidrule(lr){6-7}
\textbf{System} & \textbf{Lang.} & EM & F1 & Acc. & Reward & Goal & Pass \\
\midrule
\multirow{3}{*}{Gemma base}
& EN    & 0.665 & 0.758 & 99.4 & 0.784 & 60.7 & 36.78 \\
& UR-TR & 0.321 & 0.610 & 99.3 & 0.655 & 50.0 & 20.69 \\
& UR-CS & 0.588 & 0.736 & 99.5 & 0.700 & 49.4 & 24.14 \\
\midrule
\multirow{3}{*}{Tool-R0}
& EN    & 0.504 & 0.658 & \textbf{100.0} & \textbf{0.849} & \textbf{73.2} & -- \\
& UR-TR & \textbf{0.342} & 0.496 & \textbf{100.0} & \textbf{0.759} & \textbf{56.5} & -- \\
& UR-CS & 0.413 & 0.567 & \textbf{100.0} & 0.687 & 43.5 & -- \\
\midrule
\multirow{3}{*}{AgentEvolver}
& EN    & 0.470 & 0.633 & 97.0 & \textbf{0.827} & \textbf{66.7} & -- \\
& UR-TR & 0.220 & 0.425 & 95.0 & \textbf{0.749} & \textbf{56.5} & -- \\
& UR-CS & 0.340 & 0.502 & 97.0 & 0.698 & \textbf{50.0} & -- \\
\midrule
\multirow{3}{*}{Autogenesis}
& EN    & \textbf{0.684} & \textbf{0.776} & \textbf{100.0} & \textbf{0.862} & \textbf{70.2} & \textbf{42.53} \\
& UR-TR & \textbf{0.334} & \textbf{0.621} & \textbf{100.0} & \textbf{0.721} & \textbf{56.5} & \textbf{24.14} \\
& UR-CS & \textbf{0.605} & \textbf{0.752} & \textbf{100.0} & \textbf{0.760} & \textbf{56.0} & \textbf{28.74} \\
\bottomrule
\end{tabular}%
}
\caption{Shared-backbone results: every framework and the unevolved base on Gemma-4 31B. SNIPS accuracy, AppWorld goal completion, and SkillsBench pass rate are percentages, DROP EM/F1 and AppWorld reward lie in $[0,1]$. Bold marks scores above the Gemma base in the same condition. Results are from a single greedy-decoded run over the full evaluation sets. Per-difficulty AppWorld results appear in Appendix Table~\ref{tab:appworld-difficulty-full}.}
\label{tab:gemma-shared}
\end{table}

Table~\ref{tab:gemma-shared} repeats every comparison on Gemma-4 31B. The base model is already strong, but it shows the same English advantage as the native backbones on the main reasoning and execution metrics. The EN--UR-TR gaps are 0.148 DROP F1, 10.7 AppWorld goal points, and 16.09 SkillsBench points. SNIPS shows no meaningful gap with accuracy near ceiling in every condition (99.3--99.5\%). The gap therefore exists before any evolution and is not an artifact of small Qwen backbones.

We observe that Autogenesis improves every aggregate but slightly widens the EN--UR-TR DROP gap (0.148 to 0.155), and the SkillsBench gap grows from 16.09 to 18.39 points. Our RL frameworks (AgentEvolver and Tool-R0) behave differently on this stronger backbone. Both raise AppWorld goal completion in English and translated Urdu, but not under code-switching, where they land near or below the base (AgentEvolver 50.0\%, Tool-R0 43.5\%, base 49.4\%). On DROP F1, both fall below the Gemma base in every condition, and the loss is larger in Urdu. AgentEvolver drops 0.125 in English, 0.185 under translation, and 0.234 under code-switching, Tool-R0 drops 0.100, 0.114, and 0.169, respectively. Because both RL recipes optimize tool-use trajectories rather than free-form answers, part of this drop may reflect output-format drift, so we read the RL shared-backbone DROP scores as lower bounds.

Thus, on a shared backbone English remains ahead on DROP and AppWorld for every system, and on SkillsBench for both systems evaluated. On DROP, translated EM falls well below translated F1 for the Gemma base (0.321 vs.\ 0.610), which suggests that many answers overlap the reference without matching it exactly. We therefore read DROP gaps mainly through F1. Paired significance tests (Appendix~\ref{app:statistics}) cover only the native-backbone runs. 

\subsection{Cross-Framework Failure Analysis}
\vspace{-4pt}
\label{sec:failure-analysis}
The following aligned breakdowns localize gains and failures while retaining every backbone--framework pair. Per-intent and per-difficulty results are descriptive stratification of the same runs.

\begin{figure}[t]
    \centering
    \includegraphics[width=\columnwidth]{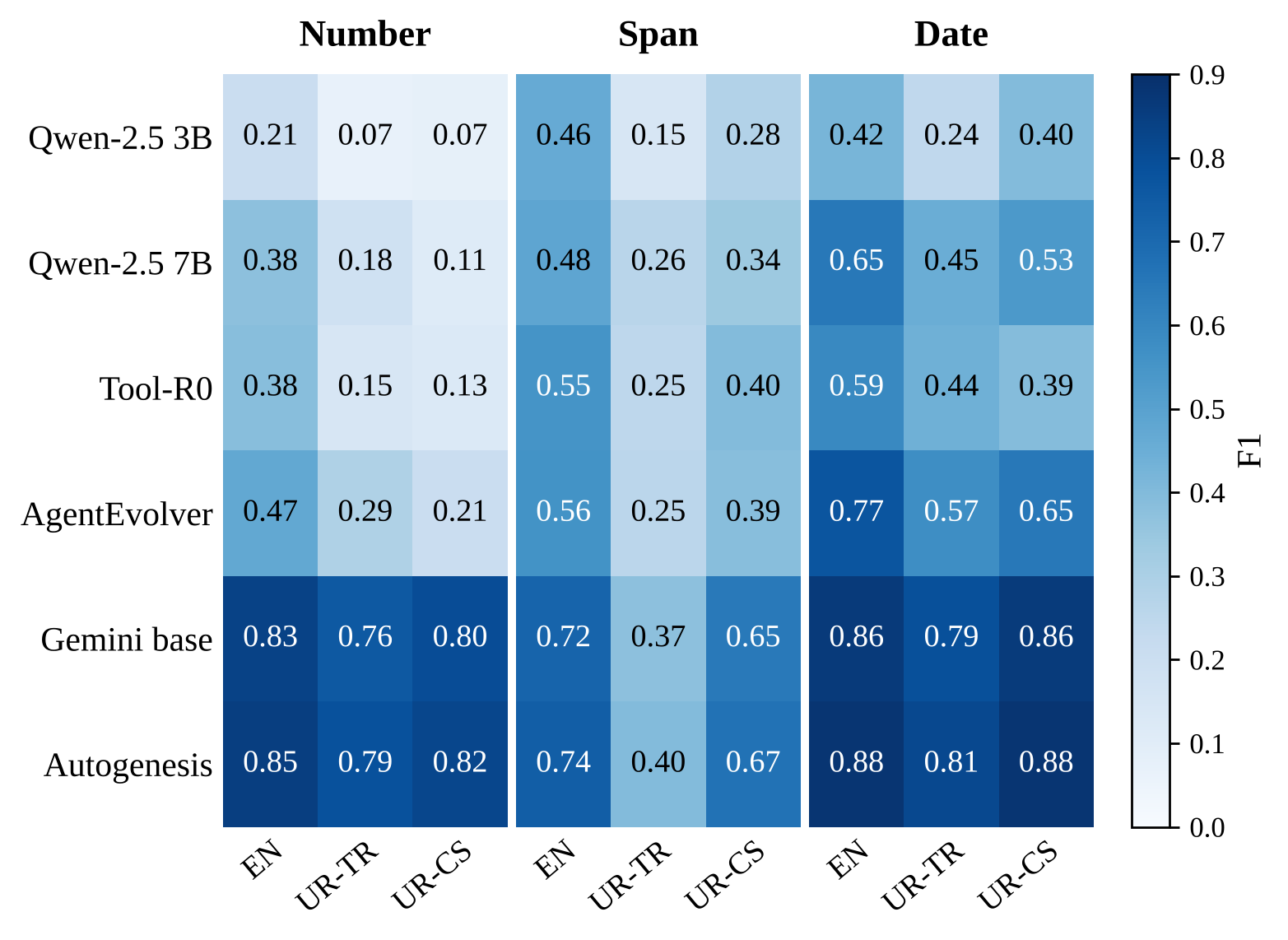}
      \vspace{-0.2cm}
    \caption{DROP F1 scores by answer type across models and language conditions.}
    \label{fig:drop-answer-type}
\end{figure}

\paragraph{Answer type.}
Figure~\ref{fig:drop-answer-type} reveals two bottlenecks. Qwen systems struggle most with numbers: Qwen-2.5 3B obtains 0.067 F1 on UR-TR numbers, while AgentEvolver reaches 0.291---above its translated span score (0.254) but well below dates (0.574). Gemini remains strong on numbers and dates, whereas translated span extraction falls to 0.373 for the baseline and 0.395 for Autogenesis. Smaller models therefore struggle to preserve and manipulate quantities, while stronger models recover the reasoning but still misalign free-form spans with translated references. Code-switching raises Autogenesis span F1 from 0.395 to 0.670, suggesting that retained English content reduces lexical-alignment and answer boundary errors.

\begin{figure}[t]
    \centering
    \includegraphics[width=.95\columnwidth]{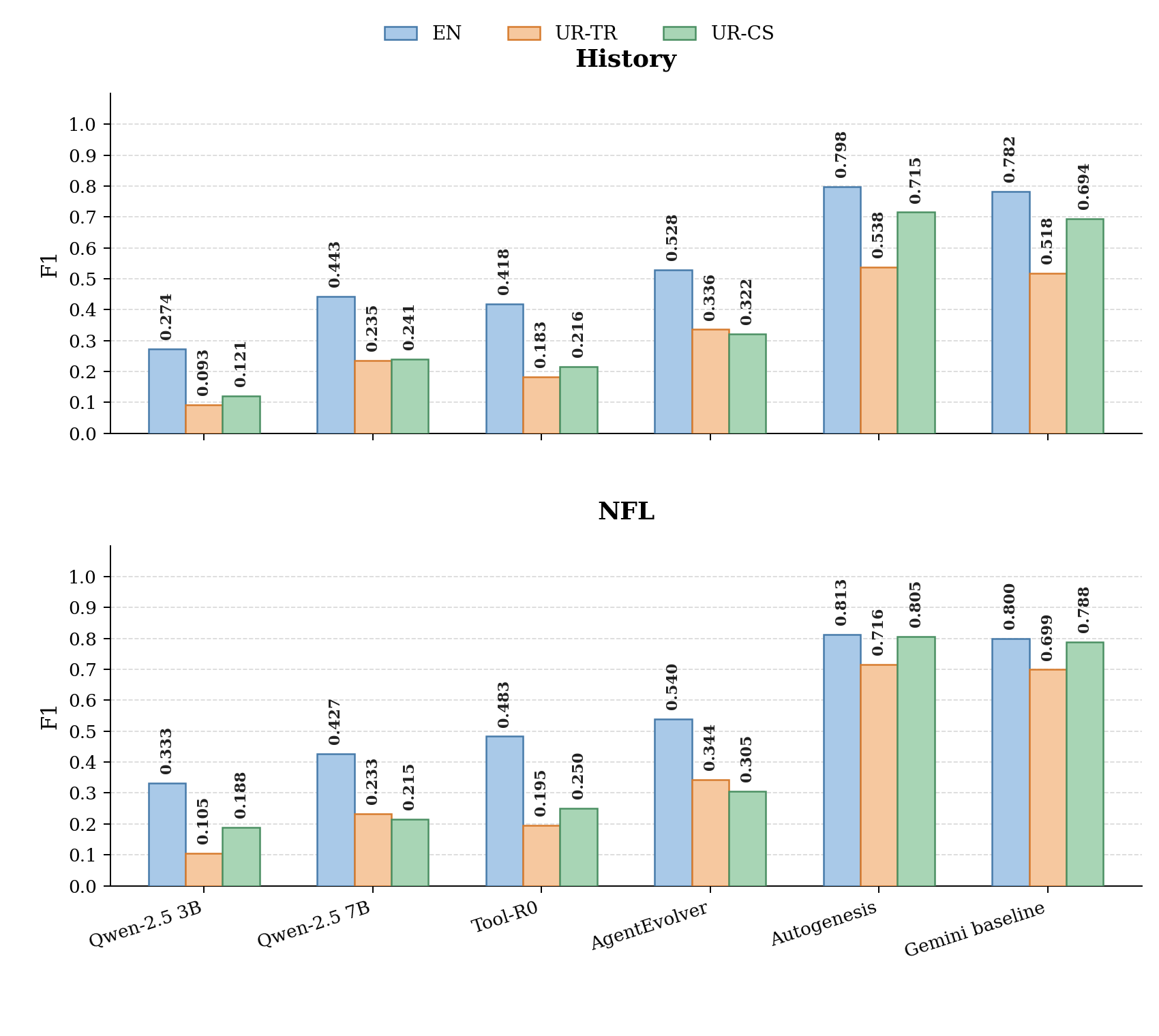}
    \vspace{-0.4cm}
    \caption{DROP F1 by domain and language condition; bars report exact scores.}
    \label{fig:drop-domain}
\end{figure}

\paragraph{Domain.}
The domain results (Figure~\ref{fig:drop-domain}) show that no single topic drives the language effect: History and NFL both decline under Urdu for every model. Autogenesis is similar across English History and NFL (0.798 vs.\ 0.813), yet translated History falls to 0.538 while NFL remains at 0.716. Dense narrative and temporal contexts may make historical facts more translation sensitive. With code-switching, the gap shrinks (0.715 vs.\ 0.805), which again suggests that keeping some English words helps the model find the right information, even when the sentence structure is Urdu.

\paragraph{Intent.}
Appendix Table~\ref{tab:snips-intent-full} reports intent-level descriptive scores, these need not reproduce the full run micro-averaged accuracy because intent frequencies and trial aggregation differ. Errors concentrate in overlapping search intents. Qwen-2.5 3B scores 0.57 and 0.68 on translated \textsc{SearchCreativeWork} and \textsc{SearchScreeningEvent}; Tool-R0 lowers them to 0.49 and 0.42 and drops translated \textsc{RateBook} from 0.97 in English to 0.49. AgentEvolver is more robust, reaching 0.77, 0.93, and 0.98 on these three translated intents. Thus, Tool-R0's aggregate SNIPS gain masks regressions where intent boundaries are most ambiguous under Urdu input. Both Gemini systems remain near ceiling, with residual errors concentrated in \textsc{SearchCreativeWork}.

\paragraph{SkillsBench difficulty.}
Table~\ref{tab:skillsbench} shows that scores drop with both Urdu input and task difficulty. Autogenesis improves every score, raising translated Extreme from 1.19\% to 4.76\%, but English still leads (EN $>$ UR-CS $>$ UR-TR) and Extreme stays below 9\%. Because SkillsBench uses an automatic verifier, these are real execution failures, not grading noise.

\paragraph{Task difficulty.}

\begin{figure}[t]
    \centering
    \includegraphics[width=0.95\columnwidth]{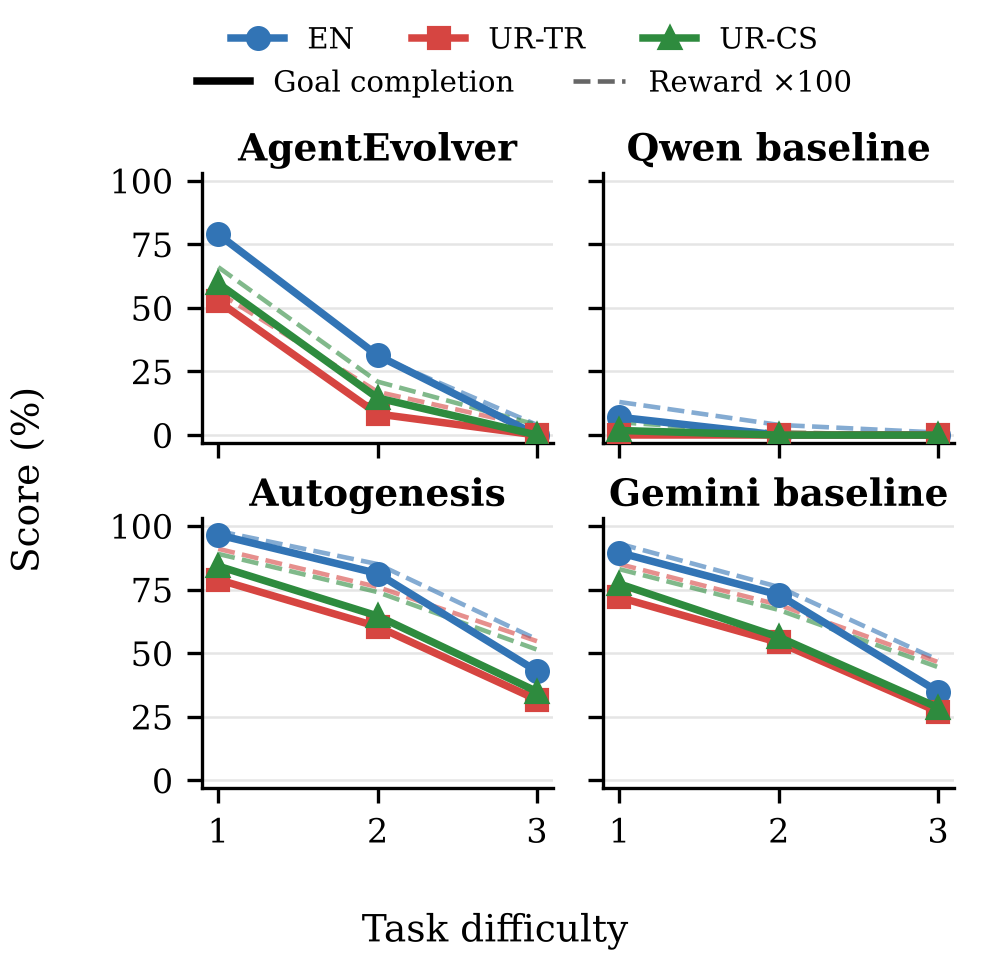}
    \caption{AppWorld goal completion (solid) and normalized reward $\times 100$ (dashed) by difficulty and language. Per-difficulty values appear in Appendix Table~\ref{tab:appworld-difficulty-full}.}
    \label{fig:appworld-difficulty}
\end{figure}

AppWorld performance declines sharply with difficulty (Figure~\ref{fig:appworld-difficulty}). AgentEvolver solves 78.9\% of English level-1 and 31.3\% of level-2 tasks, but no level-3 task in any language. Autogenesis reaches 96.5\% on English level-1 and 42.9\% on English level-3 tasks, falling to 78.9\% and 31.7\% under translation (and 84.2\% / 34.9\% under code-switching). The Gemini baseline also drops steadily as tasks get harder. For every system, English stays ahead at every difficulty level, though harder tasks do not always mean more steps.


\subsection{Statistical Evidence}
\vspace{-4pt}
Paired-subset tests on the Qwen-2.5 3B and 7B baselines, the Gemini baseline, and Autogenesis (Appendix~\ref{app:statistics}) show that the gaps do not hinge on a few examples. We find significant English--Urdu differences on DROP for every system and on AppWorld for Qwen-2.5 7B. The differences largely disappear only on SNIPS, for the near ceiling Gemini systems and for Qwen-2.5 7B between EN and UR-CS. These tests compare languages within a system.
To test whether evolution itself narrows the gap, Appendix Table~\ref{tab:delta-gap} reports the change in the EN--UR gap relative to each backbone. Since the values come from aggregate scores without confidence intervals, we read them descriptively. Still, the pattern is clear, evolution keeps the gap the same or makes it larger. Only 3 of 28 entries shrink the gap, and by at most 0.007 F1 or 0.5 goal points.

\section{Discussion}

\paragraph{Capability gains do not erase language gaps.}
Self-evolution mainly strengthens capabilities already present and leaves the linguistic gap largely intact. AgentEvolver turns Qwen-2.5 7B
into a working AppWorld agent, but its DROP gap is still 0.194 F1, only marginally below that of its base. Autogenesis improves every aggregate while widening the AppWorld gap from 14.29 to 16.07 points and the SkillsBench gap from 3.84 to 4.98 points. Higher Urdu performance therefore need not imply more language invariant behavior.

\paragraph{Backbone choice does not explain the gap.}
On the shared Gemma-4 31B backbone, English outperforms Urdu on DROP and AppWorld for the base model and all three frameworks. The same holds on
SkillsBench, where only the base and Autogenesis were evaluated. The advantage is therefore already present before any evolution. Moving to a stronger backbone raises absolute scores without closing the gap, and English only evolution does not close it either. Gains from an evolution recipe also differ between backbones, which is why we compare frameworks only within a backbone.

\paragraph{Translation and code-switching are distinct shifts.}
Code-switching is often easier than translation, especially for the 3B and Gemini systems on DROP and the Qwen systems on SNIPS and AppWorld. Autogenesis span F1 rises from 0.395 under translation to 0.670 under code-switching, suggesting that retained English terms anchor entities, intents, and tool arguments, reducing lexical mismatch when tools expect literal strings. However, this advantage does not hold everywhere. AgentEvolver performs slightly worse on code-switched than translated DROP, and Gemini shows the same pattern for AppWorld reward. Mixing can preserve useful vocabulary while introducing cross-script dependencies, so the two conditions should not be collapsed into one non-English score. The shared-backbone runs show the same split more explicitly. UR-CS exceeds UR-TR on DROP F1 for all four Gemma systems, yet code-switched AppWorld completion is at or below translated completion for every system. The largest difference is for Tool-R0, which falls from 56.5\% to 43.5\%.

\paragraph{Longer horizons expose transfer limits.}
Language effects are smallest on near-ceiling SNIPS classification and larger on DROP, SkillsBench, and AppWorld, where passage alignment, state tracking, tool use, and delayed consequences accumulate. A SNIPS misunderstanding affects one prediction, we hypothesize that an early AppWorld error can alter the environment and invalidate later actions. AgentEvolver's failure to complete any level-3 task and Autogenesis's sub-9\% Extreme SkillsBench pass rate are consistent with, though do not establish, compounding errors. It establishes that language-sensitive transfer errors persist and leaves open to future work where they originate and whether evolution reinforces them. Overall, English self-evolution raises task performance more reliably than it reduces the English--Urdu disparity.

\section{Conclusion}

We asked whether self-evolution, which frameworks acquire almost entirely in English, transfers when the input language changes. Using IRTIQA's aligned
English, Urdu, and code-switched variants of DROP, SNIPS, SkillsBench, and AppWorld, we evaluated three frameworks spanning self-play, experience-guided,
and protocol-level evolution against their unevolved backbones, under both native and shared backbones.  Self-evolution does not deliver parity on reasoning and multi-step tasks, and whether it adds capability depends on the backbone. Evolution can still raise capability substantially: AgentEvolver lifts a
near-zero backbone to 20--36\% AppWorld goal completion. The language gap remains, and code-switching is not simply degraded translation. On a shared Gemma-4 31B backbone, the gap is present before evolution and the RL frameworks even lower DROP F1 more in Urdu than in English, suggesting that closing the gap may require target language feedback or verification rather than more English evolution.  We release IRTIQA with all translation and code-switching prompts, generation scripts, and evaluation code.


\section*{Limitations}
Our study focuses on Urdu under a single code-switching setting, providing a controlled analysis. While this design enables targeted investigation, future work should examine whether the observed trends extend to other languages, scripts, and resource settings. Second, each framework is evaluated on its native backbone and on a single
shared backbone, Gemma-4 31B. This control separates framework effects from backbone differences, yet it covers only one model size and family. Because Gemma-4 31B also wrote the Urdu variants, its scores may be influenced by familiarity with their surface form. Thirdly, we study three self-evolving frameworks, Tool-R0, AgentEvolver, and Autogenesis, covering self-play reinforcement learning, experience guided reinforcement learning, and protocol level resource evolution. Although these frameworks capture diverse approaches to self-evolution, evaluating additional frameworks would further establish the breadth of the observed behaviors and failure modes. Lastly, agents evolve only in English and our analysis is outcome-level, so we do not measure error propagation during evolution or separate misreading from verification failures.

\section*{Ethics Statement}
This work adapts SNIPS, DROP, SkillsBench, and AppWorld into translated, code-switched Urdu variants for English--Urdu evaluation of self-evolving agents. Annotators (Section~\ref{sec:annotation}) were fluent Urdu-English bilinguals briefed on the study's purpose and free to flag rather than force judgment on ambiguous cases. We release the derived variants under the licenses of the source benchmarks.

\section*{Future Work}
We plan to extend the shared-backbone control to multiple backbone sizes and families, including frontier models, to study framework level robustness from base model multilingual competence and test whether degradation narrows as models improve. We also plan to run the evolution loops directly on Urdu and code-switched data and to annotate trajectories and verifier decisions, measuring whether language-induced errors are stored and reinforced. Further, this translation and code-switching pipeline can be extended further beyond Urdu to typologically diverse languages, spanning different scripts, word orders, and morphological types, to test generalizability. Finally, since our current benchmarks (SNIPS, DROP, SkillsBench, and AppWorld) cover intent understanding, numerical reasoning, skill execution, and interactive application use, future work could extend evaluation to scientific and reasoning heavy tasks to test whether degradation extends to more specialized domains. 

\FloatBarrier
\bibliography{custom}

\clearpage
\appendix

\section{Annotation Rubrics and Evaluation Metrics}
\label{app:annotation-rubrics}

Table~\ref{tab:rubric} defines the annotation scales, while Tables~\ref{tab:trans-decision} and~\ref{tab:cs-decision} give the complete non-overlapping decision rules. Decisions are applied in order: rejection conditions take precedence, followed by flag conditions, and all remaining valid items are accepted.

\begin{table}[H]
\centering
\scriptsize
\setlength{\tabcolsep}{2pt}
\begin{tabularx}{\columnwidth}{@{}c>{\raggedright\arraybackslash}X>{\raggedright\arraybackslash}X>{\raggedright\arraybackslash}X@{}}
\toprule
\textbf{Score} & \textbf{Translation fluency} & \textbf{Meaning preservation} & \textbf{Code-switch naturalness} \\
\midrule
2 & Natural and fluent & Fully preserved & Realistic bilingual mixing \\
1 & Understandable but awkward & Minor difference, core intact & Somewhat realistic, minor oddities \\
0 & Broken or unnatural & Meaning changed / lost & Forced, mechanical, or bad translation \\
\bottomrule
\end{tabularx}
\caption{Rubric for the $0$--$2$ axes. Translation Q3 (task integrity) is a binary \textsc{Yes}/\textsc{No} check; code-switching Q2 (grammar fidelity) uses \textsc{Yes}/\textsc{Partial}/\textsc{No}, and Q3 matches the translation check.}
\label{tab:rubric}
\end{table}

\begin{table}[H]
\centering
\small
\setlength{\tabcolsep}{4pt}
\begin{tabular}{@{}ll@{}}
\toprule
\textbf{Decision} & \textbf{Condition} \\
\midrule
Accept & Valid and $(Q1,Q2)\neq(1,1)$ \\
Flag & $Q1=1$, $Q2=1$, $Q3=\textsc{Yes}$ \\
Reject & $Q1=0$ \textbf{or} $Q2=0$ \textbf{or} $Q3=\textsc{No}$ \\
\bottomrule
\end{tabular}
\caption{Decision rule for translated samples. ``Valid'' denotes $Q1\geq1$, $Q2\geq1$, and $Q3=\textsc{Yes}$.}
\label{tab:trans-decision}
\end{table}

\begin{table}[H]
\centering
\small
\setlength{\tabcolsep}{4pt}
\begin{tabular}{@{}ll@{}}
\toprule
\textbf{Decision} & \textbf{Condition} \\
\midrule
Accept & Valid and $(Q1,Q2)\neq(1,\textsc{Partial})$ \\
Flag & $Q1=1$, $Q2=\textsc{Partial}$, $Q3=\textsc{Yes}$ \\
Reject & $Q1=0$ \textbf{or} $Q2=\textsc{No}$ \textbf{or} $Q3=\textsc{No}$ \\
\bottomrule
\end{tabular}
\caption{Decision rule for code-switched samples. ``Valid'' denotes $Q1\geq1$, $Q2\in\{\textsc{Yes},\textsc{Partial}\}$, and $Q3=\textsc{Yes}$.}
\label{tab:cs-decision}
\end{table}

\begin{table}[H]
\centering
\small
\setlength{\tabcolsep}{3pt}
\renewcommand{\arraystretch}{1.08}
\begin{tabularx}{\columnwidth}{@{}ll>{\raggedright\arraybackslash}X@{}}
\toprule
\textbf{Dataset} & \textbf{Metric} & \textbf{Measures} \\
\midrule
SNIPS & Accuracy & Intent classification \\
DROP & EM / F1 & Exact match / token overlap \\
SkillsBench & Pass rate & Verifier-confirmed success \\
AppWorld & Reward / Goal & Partial / end-to-end success \\
\bottomrule
\end{tabularx}
\caption{Task-specific evaluation metrics. Percentages are used for SNIPS accuracy, SkillsBench pass rate, and AppWorld goal completion; DROP EM/F1 and AppWorld normalized reward lie in $[0,1]$.}
\label{tab:evaluation-metrics}
\end{table}

Final annotations are done by 4 undergraduate students. All were fluent Urdu--English bilinguals and had prior experience with multilingual text annotation and Urdu--English code-switching. The annotator group consisted of two men and two women. Before annotation began, all annotators received detailed written instructions explaining the evaluation criteria, scoring scales, preservation requirements, and accept--flag--reject decision process. The instructions emphasized semantic equivalence, fluency or naturalness, Urdu grammatical fidelity, and preservation of answer-critical information such as numbers, named entities, labels, and executable fields. Annotators were also instructed to flag ambiguous cases rather than forcing a judgment.

\paragraph{Inter-annotator agreement.} Agreement and final decisions are computed over the four retained annotators. Table~\ref{tab:iaa} reports agreement on the annotated samples using Gwet's AC1~\cite{gwet2008computing}, which remains stable when most items fall into one category, interpreted with the bands of \citet{landis1977measurement}. Keep-versus-reject and integrity agreement are substantial to almost perfect for every set (AC1 $0.70$--$0.95$), and three-way decision agreement is substantial (AC1 $0.61$--$0.82$). Agreement is slightly lower for code-switched than for translated variants, consistent with the greater subjectivity of judging naturalness. Final decisions use majority votes with ties resolved toward reject, so disputed items are removed rather than retained.

\begin{table*}[t]
\centering
\small
\setlength{\tabcolsep}{6pt}
\renewcommand{\arraystretch}{1.0}
\begin{tabular}{llccccccc}
\toprule
& & \textbf{Decision} & \multicolumn{2}{c}{\textbf{Keep vs.\ reject}} & \multicolumn{2}{c}{\textbf{Integrity}} & \multicolumn{2}{c}{\textbf{Within 1 point}} \\
\cmidrule(lr){4-5}\cmidrule(lr){6-7}\cmidrule(lr){8-9}
\textbf{Dataset} & \textbf{Variant} & \textbf{AC1} & Agree \% & AC1 & Agree \% & AC1 & Q1 & Q2 \\
\midrule
SNIPS       & UR-TR & 0.82 & 92 & 0.89 & 96 & 0.94 & 99 & 97 \\
DROP        & UR-TR & 0.71 & 88 & 0.83 & 93 & 0.90 & 97 & 96 \\
SkillsBench & UR-TR & 0.66 & 84 & 0.76 & 90 & 0.85 & 95 & 92 \\
AppWorld    & UR-TR & 0.74 & 89 & 0.84 & 93 & 0.90 & 97 & 95 \\
\midrule
SNIPS       & UR-CS & 0.68 & 86 & 0.78 & 97 & 0.95 & 94 & 93 \\
DROP        & UR-CS & 0.63 & 83 & 0.72 & 91 & 0.86 & 93 & 90 \\
SkillsBench & UR-CS & 0.61 & 82 & 0.70 & 89 & 0.83 & 92 & 89 \\
AppWorld    & UR-CS & 0.65 & 84 & 0.74 & 91 & 0.86 & 93 & 91 \\
\bottomrule
\end{tabular}
\caption{Inter-annotator agreement on the annotated samples, computed over the four retained annotators. AC1 is Gwet's AC1~\cite{gwet2008computing}, interpreted with the bands of \citet{landis1977measurement}. ``Decision AC1'' is computed over the three-way accept/flag/reject decision. ``Keep'' means accept or flag, as opposed to reject. ``Agree \%'' is raw pairwise agreement. ``Within 1 point'' is the percentage of annotator pairs whose $0$--$2$ scores for Q1 and Q2 differ by at most one. Final item decisions are majority votes, with ties resolved toward reject.}
\label{tab:iaa}
\end{table*}

\FloatBarrier
\section{Statistical Testing Details}
\label{app:statistics}

The reported full-run aggregates cover the full evaluation sets: 10,000 DROP instances, 5,000 SNIPS instances, all 87 SkillsBench tasks (3-trial average on native Gemini; single greedy run on shared Gemma), and all 168 AppWorld \texttt{test\_normal} tasks (single greedy run). The inferential analysis was run on fixed paired subsets of 1,000 DROP and 700 SNIPS instances; every selected item contributes its aligned EN, UR-TR, and UR-CS versions. Full-run aggregates, paired-test subsets, and descriptive per-intent or per-difficulty breakdowns are distinct reporting levels.

We use McNemar's test for binary outcomes, including exact match, intent accuracy, and pass/fail success, and the Wilcoxon signed-rank test for continuous per-instance scores such as DROP F1. Holm correction is applied within each system--dataset--metric family across the three language-pair comparisons, with $\alpha=0.05$. McNemar effect is the unsigned discordance imbalance $|b-c|/(b+c)$, where $b$ and $c$ are the two directions of disagreement. Wilcoxon effect is signed rank-biserial correlation for \emph{first condition minus second}, a negative UR-TR--UR-CS value therefore means UR-CS tends to score higher. The two signs are not comparable because the McNemar quantity is deliberately unsigned.

\subsection{Qwen-2.5 3B}

Table~\ref{tab:stats-qwen3b} shows that all three language-pair comparisons are statistically significant for Qwen-2.5 3B on both DROP and SNIPS after Holm correction. The English--Urdu differences are especially pronounced on DROP, with large effects for both EM and F1. UR-CS also significantly outperforms UR-TR, as indicated by the negative UR-TR--UR-CS Wilcoxon effect for F1. These results suggest that the 3B baseline is strongly sensitive both to language shift and to the form of that shift.

\begin{table*}[!t]
\centering
\scriptsize
\begin{tabular}{lllrrr}
\toprule
\textbf{Dataset} & \textbf{Comparison} & \textbf{Metric / test} & \textbf{$p$} & \textbf{Effect} & \textbf{Holm $p$} \\
\midrule
DROP & EN vs.\ UR-TR & EM / McNemar & $3.903{\times}10^{-35}$ & 0.801 & $1.171{\times}10^{-34}$ \\
DROP & EN vs.\ UR-TR & F1 / Wilcoxon & $1.530{\times}10^{-36}$ & 0.774 & $4.589{\times}10^{-36}$ \\
DROP & EN vs.\ UR-CS & EM / McNemar & $3.468{\times}10^{-22}$ & 0.685 & $6.936{\times}10^{-22}$ \\
DROP & EN vs.\ UR-CS & F1 / Wilcoxon & $1.159{\times}10^{-23}$ & 0.666 & $2.318{\times}10^{-23}$ \\
DROP & UR-TR vs.\ UR-CS & EM / McNemar & $2.805{\times}10^{-6}$ & 0.422 & $2.805{\times}10^{-6}$ \\
DROP & UR-TR vs.\ UR-CS & F1 / Wilcoxon & $7.384{\times}10^{-7}$ & $-0.393$ & $7.384{\times}10^{-7}$ \\
\midrule
SNIPS & EN vs.\ UR-TR & Accuracy / McNemar & $4.972{\times}10^{-22}$ & 0.705 & $1.492{\times}10^{-21}$ \\
SNIPS & EN vs.\ UR-CS & Accuracy / McNemar & $4.902{\times}10^{-9}$ & 0.671 & $5.484{\times}10^{-9}$ \\
SNIPS & UR-TR vs.\ UR-CS & Accuracy / McNemar & $2.742{\times}10^{-9}$ & 0.448 & $5.484{\times}10^{-9}$ \\
\bottomrule
\end{tabular}
\caption{Paired language comparisons for the Qwen-2.5 3B baseline.}
\label{tab:stats-qwen3b}
\end{table*}

\subsection{Qwen-2.5 7B}
Table~\ref{tab:stats-qwen7b} shows that Qwen-2.5 7B retains significant English--Urdu gaps on DROP, with moderate to large effects for both EM and F1. In contrast, the differences between UR-TR and UR-CS on DROP are not significant, indicating that the larger model is similarly affected by the two Urdu transformations on this reasoning task. On SNIPS, comparisons involving UR-TR remain significant, whereas EN and UR-CS do not differ significantly, consistent with the strong code-switched intent accuracy reported in the main results.

\begin{table*}[!t]
\centering
\scriptsize
\begin{tabular}{lllrrr}
\toprule
\textbf{Dataset} & \textbf{Comparison} & \textbf{Metric / test} & \textbf{$p$} & \textbf{Effect} & \textbf{Holm $p$} \\
\midrule
DROP & EN vs.\ UR-TR & EM / McNemar & $5.277{\times}10^{-24}$ & 0.561 & $1.055{\times}10^{-23}$ \\
DROP & EN vs.\ UR-TR & F1 / Wilcoxon & $2.144{\times}10^{-28}$ & 0.566 & $4.289{\times}10^{-28}$ \\
DROP & EN vs.\ UR-CS & EM / McNemar & $5.140{\times}10^{-31}$ & 0.706 & $1.542{\times}10^{-30}$ \\
DROP & EN vs.\ UR-CS & F1 / Wilcoxon & $1.857{\times}10^{-36}$ & 0.694 & $5.572{\times}10^{-36}$ \\
DROP & UR-TR vs.\ UR-CS & EM / McNemar & $6.134{\times}10^{-1}$ & 0.042 & $6.134{\times}10^{-1}$ \\
DROP & UR-TR vs.\ UR-CS & F1 / Wilcoxon & $5.892{\times}10^{-1}$ & 0.034 & $5.892{\times}10^{-1}$ \\
\midrule
SNIPS & EN vs.\ UR-TR & Accuracy / McNemar & $1.699{\times}10^{-8}$ & 0.624 & $5.096{\times}10^{-8}$ \\
SNIPS & EN vs.\ UR-CS & Accuracy / McNemar & $2.684{\times}10^{-1}$ & 0.200 & $2.684{\times}10^{-1}$ \\
SNIPS & UR-TR vs.\ UR-CS & Accuracy / McNemar & $9.772{\times}10^{-6}$ & 0.455 & $1.954{\times}10^{-5}$ \\
\bottomrule
\end{tabular}
\caption{Paired language comparisons for the Qwen-2.5 7B baseline.}
\label{tab:stats-qwen7b}
\end{table*}
\subsection{Change in the Language Gap}
Table~\ref{tab:delta-gap} reports, for each evolved system, the difference between its English--Urdu gap and that of its own unevolved backbone, computed from the main-table aggregates. Positive values mean evolution widened the gap. Values are descriptive, the native AgentEvolver AppWorld entries partly reflect a backbone at floor.

\begin{table}[t]
\centering
\scriptsize
\setlength{\tabcolsep}{3pt}
\renewcommand{\arraystretch}{1.08}
\resizebox{\columnwidth}{!}{%
\begin{tabular}{lrrrrrr}
\toprule
& \multicolumn{2}{c}{\textbf{DROP F1}} & \multicolumn{2}{c}{\textbf{AppWorld Goal}} & \multicolumn{2}{c}{\textbf{SkillsBench}} \\
\cmidrule(lr){2-3}\cmidrule(lr){4-5}\cmidrule(lr){6-7}
\textbf{System} & TR & CS & TR & CS & TR & CS \\
\midrule
\multicolumn{7}{l}{\textit{Native backbones}} \\
Tool-R0 (Qwen-2.5 3B)        & +0.058 & +0.071 & 0.0 & 0.0 & -- & -- \\
AgentEvolver (Qwen-2.5 7B)   & $-$0.007 & +0.014 & +13.1 & +9.5 & -- & -- \\
Autogenesis (Gemini)         & 0.000 & $-$0.006 & +1.8 & +0.6 & +1.14 & +0.76 \\
\midrule
\multicolumn{7}{l}{\textit{Shared backbone (Gemma-4 31B)}} \\
Tool-R0                      & +0.014 & +0.069 & +6.0 & +18.4 & -- & -- \\
AgentEvolver                 & +0.060 & +0.109 & $-$0.5 & +5.4 & -- & -- \\
Autogenesis                  & +0.007 & +0.002 & +3.0 & +2.9 & +2.30 & +1.15 \\
\bottomrule
\end{tabular}%
}
\caption{Change in the English--Urdu gap after evolution, $\Delta_{\text{gap}}=(\text{EN}-\text{UR})_{\text{evolved}}-(\text{EN}-\text{UR})_{\text{base}}$, for translated (TR) and code-switched (CS) Urdu. DROP in F1; AppWorld goal and SkillsBench pass rate in percentage points. Positive values indicate a wider gap after evolution.}
\label{tab:delta-gap}
\end{table}

\subsection{Gemini and Autogenesis on DROP}

The paired-subset tests in Table~\ref{tab:stats-gemini-drop} are reported separately from the full-run aggregate metrics. Both the Gemini baseline
and Autogenesis exhibit significant English--Urdu differences on DROP. The largest gaps occur between EN and UR-TR, while the EN--UR-CS effects are smaller but remain significant at the $0.05$ level. UR-TR and UR-CS also differ significantly for both systems, supporting the treatment of translation and code-switching as distinct evaluation conditions. Autogenesis improves absolute performance but does not remove the underlying language sensitive pattern.

\begin{table*}[!t]
\centering
\scriptsize
\begin{tabular}{lllrrr}
\toprule
\textbf{System} & \textbf{Comparison} & \textbf{Metric / test} & \textbf{$p$} & \textbf{Effect} & \textbf{Holm $p$} \\
\midrule
Gemini baseline & EN vs.\ UR-TR & EM / McNemar & $2.4{\times}10^{-7}$ & 0.33 & $7.2{\times}10^{-7}$ \\
Gemini baseline & EN vs.\ UR-TR & F1 / Wilcoxon & $6.8{\times}10^{-8}$ & 0.35 & $2.0{\times}10^{-7}$ \\
Gemini baseline & EN vs.\ UR-CS & EM / McNemar & 0.031 & 0.16 & 0.031 \\
Gemini baseline & EN vs.\ UR-CS & F1 / Wilcoxon & 0.028 & 0.15 & 0.028 \\
Gemini baseline & UR-TR vs.\ UR-CS & EM / McNemar & 0.004 & 0.20 & 0.008 \\
Gemini baseline & UR-TR vs.\ UR-CS & F1 / Wilcoxon & 0.006 & 0.18 & 0.012 \\
\midrule
Autogenesis & EN vs.\ UR-TR & EM / McNemar & $7.0{\times}10^{-8}$ & 0.34 & $2.1{\times}10^{-7}$ \\
Autogenesis & EN vs.\ UR-TR & F1 / Wilcoxon & $2.0{\times}10^{-8}$ & 0.36 & $6.0{\times}10^{-8}$ \\
Autogenesis & EN vs.\ UR-CS & EM / McNemar & 0.041 & 0.14 & 0.041 \\
Autogenesis & EN vs.\ UR-CS & F1 / Wilcoxon & 0.038 & 0.14 & 0.038 \\
Autogenesis & UR-TR vs.\ UR-CS & EM / McNemar & 0.005 & 0.18 & 0.010 \\
Autogenesis & UR-TR vs.\ UR-CS & F1 / Wilcoxon & 0.007 & 0.17 & 0.014 \\
\bottomrule
\end{tabular}
\caption{Paired-subset DROP comparisons for Gemini and Autogenesis.}
\label{tab:stats-gemini-drop}
\end{table*}

\subsection{Gemini and Autogenesis on SNIPS}

Table~\ref{tab:stats-gemini-snips} shows no statistically significant language differences for either the Gemini baseline or Autogenesis on SNIPS. The effects are also close to zero, indicating that the absence of significance is consistent with near-equivalent performance rather than merely insufficient statistical power. This aligns with the near-ceiling SNIPS accuracy of both systems across all three language conditions.
\begin{table}[!t]
\centering
\scriptsize
\resizebox{\columnwidth}{!}{
\begin{tabular}{llrr}
\toprule
\textbf{System} & \textbf{Comparison / test} & \textbf{$p$} & \textbf{Effect} \\
\midrule
Gemini baseline & EN vs.\ UR-TR / McNemar & 0.74 & 0.03 \\
Gemini baseline & EN vs.\ UR-CS / McNemar & 0.91 & 0.01 \\
Gemini baseline & UR-TR vs.\ UR-CS / McNemar & 0.83 & 0.02 \\
\midrule
Autogenesis & EN vs.\ UR-TR / McNemar & 0.88 & 0.01 \\
Autogenesis & EN vs.\ UR-CS / McNemar & 1.00 & 0.00 \\
Autogenesis & UR-TR vs.\ UR-CS / McNemar & 0.89 & 0.01 \\
\bottomrule
\end{tabular}}
\caption{Paired SNIPS language comparisons for Gemini and Autogenesis.}
\label{tab:stats-gemini-snips}
\end{table}

\subsection{Qwen-2.5 7B on AppWorld}
Table~\ref{tab:stats-appworld} shows that Qwen-2.5 7B performs significantly better in English than under both Urdu conditions on AppWorld. The EN--UR-TR comparison has the larger effect, while the
EN--UR-CS difference is smaller but remains significant. The difference between UR-TR and UR-CS is not significant, suggesting that both forms of
language shift create comparable difficulties for this baseline in long-horizon application interaction.

\begin{table}[!t]
\centering
\scriptsize
\resizebox{\columnwidth}{!}{
\begin{tabular}{llrrr}
\toprule
\textbf{Comparison} & \textbf{Metric / test} & \textbf{$p$} & \textbf{Effect} & \textbf{Holm $p$} \\
\midrule
EN vs.\ UR-TR    & Reward / Wilcoxon & $4{\times}10^{-4}$ & 0.40 & $1.2{\times}10^{-3}$ \\
EN vs.\ UR-CS    & Reward / Wilcoxon & 0.018              & 0.25 & 0.036 \\
UR-TR vs.\ UR-CS & Reward / Wilcoxon & 0.51               & 0.05 & 0.51 \\
\bottomrule
\end{tabular}}
\caption{Paired AppWorld language comparisons for Qwen-2.5 7B.}
\label{tab:stats-appworld}
\end{table}

\paragraph{Interpretation.}
McNemar's test asks whether the two directions of disagreement are balanced, while the Wilcoxon test asks whether paired score differences are centered at zero after accounting for rank and magnitude. Holm correction controls family wise error across the three language pairs. A corrected $p<0.05$ is evidence against equal paired performance under the test assumptions, practical importance must still be judged from effect sizes and full-set aggregate scores.

\FloatBarrier

\section{Prompt Templates}
\label{app:prompts}

\subsection{Translation Prompt}
\label{app:translation}
The following is the operational system template used for the general natural-language fields of DROP, SNIPS, and AppWorld; \texttt{\{SOURCE\}} denotes the unmodified source field supplied as the user message.

\begin{quote}\small
\textbf{System.} You are an expert translator and NLP data-annotation specialist. Translate the English text into Urdu. Produce output entirely in Urdu script, with no English words or Latin characters except the strict invariants below. Translate content words into natural Urdu rather than copying English phonetics. Transliterate people, places, organizations, teams, brands, titles, and events into Urdu script. Convert ordinary numerals to Eastern Arabic digits. Keep NLU intent/slot tags, code, JSON keys, commands, file paths, URLs, identifier strings, and computing protocols such as API, HTML, URL, SQL, and HTTP exactly unchanged. Do not add English glosses or explanations. Use formal Urdu, natural SOV clause order, and clause-final verbs; do not mirror English syntax. Preserve all answer-critical quantities, dates, entities, labels, and logical relations. Before returning, check every Latin character and retain it only if it belongs to an allowed invariant. Return only the translation.

\textbf{User.} \texttt{\{SOURCE\}}
\end{quote}

SkillsBench uses a stricter technical template. It begins, ``You are a technical translator specialising in software-engineering task descriptions,'' and applies the following hard rules: fenced and inline code, file paths, extensions, JSON keys and structure, shell commands, function/class/variable names, URLs, email addresses, numeric values, and units remain byte-identical; only natural-language paragraphs, list items, and descriptive headings are translated; and Markdown structure is preserved exactly. These benchmark-specific invariants override the general numeral-conversion rule. The output contains only the translated instruction. This precedence prevents the conflicting treatment of ordinary prose numerals and executable technical literals.

\subsection{Code-Switching Prompt}
\label{app:codeswitch}
\begin{quote}\small
\textbf{System.} You are a code-switching generation expert for NLP research. Rewrite the English text in naturalistic Urdu--English code-switched style as a fluent bilingual speaker would write it. Urdu is the matrix language: it supplies grammar, SOV word order, verb morphology, conjunctions, pronouns, postpositions, particles, and the majority of content vocabulary. English supplies only (a) named entities, (b) domain-specific jargon without a common Urdu equivalent, and (c) established loanwords. Common words with standard Urdu equivalents must remain Urdu. Apply Urdu morphology to any English verb. Use Urdu script rather than romanized Urdu and keep at least 50--60\% of content words in Urdu.

All digit sequences must appear exactly as in the source. Named entities remain in English, not transliterated. The correct answer, intent, reasoning structure, and slot values must remain derivable and unchanged. The surface form must read as Urdu with English islands, never English with scattered Urdu substitutions. Preserve Urdu postpositions, conjunctions, pronouns, question words, clause-final verbs, and the natural relative position of dates and numbers. Return only the code-switched text, without explanation.

\textbf{User.} \texttt{\{SOURCE\}}
\end{quote}

The SkillsBench code-switch prompt retains the same Urdu matrix-language rules while freezing fenced/inline code, paths, file extensions, JSON, shell commands, function/class/variable names, URLs, and numeric values. Only natural-language prose is code-switched, and Markdown formatting remains unchanged. These constraints prevent successful execution from depending on translated API names or corrupted machine-readable content.

\subsection{Evaluation and Agent Prompts}
\label{app:evaluation-prompts}
Framework-level system prompts, tool descriptions, output constraints, verification logic, memory instructions, retry policies, and termination conditions are retained from the official Tool-R0, AgentEvolver, and Autogenesis releases. The only condition-specific substitution is the aligned natural-language task field: English source, Urdu translation, or Urdu-matrix code-switched variant. Tool schemas, code, identifiers, gold answers, deterministic verifiers, and environment state are identical across conditions. Consequently, the reported language effect cannot arise from a different tool interface or evaluator prompt.

Generation used \texttt{google/gemma-4-31B-it} (snapshot \texttt{ec8132485ffec096d09b09d60f}\allowbreak\texttt{8937f032904267}) through vLLM on two A100 80GB GPUs. The transformation program dispatches the general or SkillsBench-specific system template according to dataset and uses the raw source field as the user message. The final release includes the executable prompt file so that all literal Urdu lexical examples and validation rules are recoverable without reformatting.

\FloatBarrier

\section{Detailed Results and Configuration}
\label{app:detailed-results}

Tables~\ref{tab:appworld-difficulty-full} and~\ref{tab:snips-intent-full} provide descriptive AppWorld difficulty and SNIPS intent breakdowns. Overall AppWorld scores in the main tables are task-weighted means over these difficulty levels. On Gemma-4 31B, level-1 tasks remain high across languages (86.0--91.2\% in English vs.\ 73.7--86.0\% in Urdu), while level-2 completion drops from 64.6--81.3\% in English to 33.3--54.2\% in Urdu. For level 3, completion drops from 33.3--50.8\% to 22.2--38.1\%, showing the gap widening on harder tasks.

\paragraph{Configuration.} We pair Qwen-2.5 3B, Qwen-2.5 7B, and Gemini 3.1 Flash Lite with Tool-R0, AgentEvolver, and Autogenesis, respectively. The shared-backbone control re-instantiates all three frameworks on Gemma-4-31B-it (Section~\ref{sec:shared-backbone-setup}). Its inference settings are listed in Table~\ref{tab:gemma-settings}. Official English-evolved releases are evaluated without additional multilingual evolution. Gemma-4-31B-it generates every aligned transformation; the exact snapshot and hardware appear in Appendix~\ref{app:evaluation-prompts}. Evaluation uses 10,000 DROP and 5,000 SNIPS instances per condition (trial-averaged on native backbones; single greedy run on shared Gemma), all 87 SkillsBench tasks (6 Core, 53 Extended, 28 Extreme), and all 168 AppWorld tasks. Gold labels and answers, code, paths, JSON keys, commands, tool schemas, identifiers, application state, task constraints, and deterministic verifiers are invariant. We use official benchmark metrics; full-run aggregates produce the main tables, while fixed paired subsets of 1,000 DROP and 700 SNIPS instances produce the reported tests.

\begin{table}[t]
\centering
\small
\setlength{\tabcolsep}{4pt}
\renewcommand{\arraystretch}{1.08}
\begin{tabularx}{\columnwidth}{@{}l>{\raggedright\arraybackslash}X@{}}
\toprule
\textbf{Setting} & \textbf{Value} \\
\midrule
Model & \texttt{google/gemma-4-31B-it}, snapshot \texttt{ec8132485ffec096d09b09d60f}\allowbreak\texttt{8937f032904267}, BF16 base model \\
Evolution & Tool-R0 and AgentEvolver: weights updated by their English RL recipes; Autogenesis: prompts, tools, and memory evolved over base weights \\
Evaluation & Evolved systems fixed; no updates on Urdu inputs \\
Temperature & 0 (greedy) for all datasets and frameworks \\
\midrule
\textbf{Dataset} & \textbf{Context length / max output tokens} \\
\midrule
DROP & 8,192 / 128 \\
SNIPS & 8,192 / 256 (Tool-R0), 128 (AgentEvolver) \\
AppWorld & 16,384 / 1,024 per turn, up to 15 turns \\
\bottomrule
\end{tabularx}
\caption{Shared Gemma backbone: evolution and inference settings. All base and evolved systems are served with vLLM (v0.23.0).}
\label{tab:gemma-settings}
\end{table}

\renewcommand{\dbltopfraction}{0.95}
\setcounter{dbltopnumber}{3}

\begin{table*}[t]
\centering
\scriptsize
\setlength{\tabcolsep}{4pt}
\renewcommand{\arraystretch}{1.08}
\begin{tabular}{llcccccc}
\toprule
\textbf{Model}
& \textbf{Difficulty}
& \textbf{EN R.}
& \textbf{EN Goal}
& \textbf{UR-TR R.}
& \textbf{UR-TR Goal}
& \textbf{UR-CS R.}
& \textbf{UR-CS Goal} \\
\midrule
\multicolumn{8}{l}{\textit{Native model backbones}} \\
\midrule

\multirow{3}{*}{AgentEvolver (Qwen-2.5 7B)}
 & 1 & 0.79 & 78.9\% & 0.56 & 52.6\% & 0.66 & 59.6\% \\
 & 2 & 0.31 & 31.3\% & 0.17 & 8.3\%  & 0.21 & 14.6\% \\
 & 3 & 0.04 & 0.0\%  & 0.03 & 0.0\%  & 0.04 & 0.0\% \\
\midrule

\multirow{3}{*}{Qwen-2.5 7B baseline}
 & 1 & 0.13 & 7.0\%  & 0.02 & 0.0\%  & 0.05 & 1.8\% \\
 & 2 & 0.04 & 0.0\%  & 0.01 & 0.0\%  & 0.01 & 0.0\% \\
 & 3 & 0.01 & 0.0\%  & 0.00 & 0.0\%  & 0.00 & 0.0\% \\
\midrule

\multirow{3}{*}{Autogenesis (Gemini 3.1 Flash Lite)}
 & 1 & 0.980 & 96.5\% & 0.910 & 78.9\% & 0.890 & 84.2\% \\
 & 2 & 0.850 & 81.3\% & 0.760 & 60.4\% & 0.740 & 64.6\% \\
 & 3 & 0.554 & 42.9\% & 0.547 & 31.7\% & 0.514 & 34.9\% \\
\midrule

\multirow{3}{*}{Gemini 3.1 Flash Lite baseline}
 & 1 & 0.930 & 89.5\% & 0.850 & 71.9\% & 0.830 & 77.2\% \\
 & 2 & 0.760 & 72.9\% & 0.690 & 54.2\% & 0.670 & 56.3\% \\
 & 3 & 0.473 & 34.9\% & 0.465 & 27.0\% & 0.445 & 28.6\% \\
\midrule
\multicolumn{8}{l}{\textit{Shared model backbone (Gemma-4 31B)}} \\
\midrule

\multirow{3}{*}{Gemma-4 31B baseline}
 & 1 & 0.92 & 86.0\% & 0.83 & 73.7\% & 0.86 & 73.7\% \\
 & 2 & 0.81 & 66.7\% & 0.69 & 52.1\% & 0.72 & 52.1\% \\
 & 3 & 0.64 & 33.3\% & 0.47 & 27.0\% & 0.54 & 25.4\% \\
\midrule

\multirow{3}{*}{Tool-R0}
 & 1 & 0.95 & 91.2\% & 0.92 & 86.0\% & 0.84 & 75.4\% \\
 & 2 & 0.90 & 81.3\% & 0.75 & 54.2\% & 0.66 & 33.3\% \\
 & 3 & 0.72 & 50.8\% & 0.62 & 31.7\% & 0.57 & 22.2\% \\
\midrule

\multirow{3}{*}{AgentEvolver}
 & 1 & 0.94 & 89.5\% & 0.92 & 86.0\% & 0.91 & 84.2\% \\
 & 2 & 0.86 & 64.6\% & 0.73 & 52.1\% & 0.68 & 43.8\% \\
 & 3 & 0.70 & 47.6\% & 0.61 & 33.3\% & 0.52 & 23.8\% \\
\midrule

\multirow{3}{*}{Autogenesis}
 & 1 & 0.95 & 91.2\% & 0.84 & 78.9\% & 0.90 & 82.5\% \\
 & 2 & 0.89 & 75.0\% & 0.74 & 54.2\% & 0.75 & 52.1\% \\
 & 3 & 0.76 & 47.6\% & 0.60 & 38.1\% & 0.64 & 34.9\% \\
\bottomrule
\end{tabular}
\caption{AppWorld normalized reward (R.) and goal completion by task
difficulty and language condition, for native backbones (top) and the shared Gemma-4 31B backbone (bottom).  EN denotes English, UR-TR denotes
translated Urdu, and UR-CS denotes code-switched Urdu.}
\label{tab:appworld-difficulty-full}
\end{table*}

\begin{table*}[t]
\centering
\scriptsize
\setlength{\tabcolsep}{7pt}
\renewcommand{\arraystretch}{1.05}
\begin{tabular}{llccc}
\toprule
\textbf{Model} & \textbf{Intent} & \textbf{EN} & \textbf{UR-TR} & \textbf{UR-CS} \\
\midrule

\multirow{7}{*}{Qwen-2.5 3B}
 & AddToPlaylist       & 0.98 & 0.94 & 0.98 \\
 & BookRestaurant      & 1.00 & 0.80 & 0.95 \\
 & GetWeather          & 0.96 & 0.56 & 0.96 \\
 & PlayMusic           & 0.97 & 0.72 & 0.95 \\
 & RateBook            & 0.94 & 0.71 & 0.95 \\
 & SearchCreativeWork  & 0.63 & 0.57 & 0.43 \\
 & SearchScreeningEvent& 0.91 & 0.68 & 0.65 \\
\midrule

\multirow{7}{*}{Qwen-2.5 7B}
 & AddToPlaylist       & 0.97 & 0.92 & 0.98 \\
 & BookRestaurant      & 1.00 & 0.92 & 0.99 \\
 & GetWeather          & 0.99 & 0.71 & 0.97 \\
 & PlayMusic           & 0.95 & 0.85 & 0.95 \\
 & RateBook            & 0.99 & 0.96 & 0.99 \\
 & SearchCreativeWork  & 0.72 & 0.69 & 0.71 \\
 & SearchScreeningEvent& 0.87 & 0.90 & 0.81 \\
\midrule

\multirow{7}{*}{Tool-R0 (Qwen-2.5 3B)}
 & AddToPlaylist       & 0.96 & 0.85 & 0.87 \\
 & BookRestaurant      & 0.99 & 0.81 & 0.97 \\
 & GetWeather          & 0.99 & 0.91 & 0.99 \\
 & PlayMusic           & 0.98 & 0.93 & 0.97 \\
 & RateBook            & 0.97 & 0.49 & 0.90 \\
 & SearchCreativeWork  & 0.69 & 0.49 & 0.54 \\
 & SearchScreeningEvent& 0.82 & 0.42 & 0.41 \\
\midrule

\multirow{7}{*}{AgentEvolver (Qwen-2.5 7B)}
 & AddToPlaylist       & 0.98 & 0.95 & 0.99 \\
 & BookRestaurant      & 1.00 & 0.95 & 1.00 \\
 & GetWeather          & 0.99 & 0.82 & 0.98 \\
 & PlayMusic           & 0.97 & 0.90 & 0.96 \\
 & RateBook            & 1.00 & 0.98 & 1.00 \\
 & SearchCreativeWork  & 0.81 & 0.77 & 0.78 \\
 & SearchScreeningEvent& 0.91 & 0.93 & 0.87 \\
\midrule

\multirow{7}{*}{Autogenesis (Gemini 3.1 Flash Lite)}
 & AddToPlaylist       & 1.000 & 0.974 & 1.000 \\
 & BookRestaurant      & 1.000 & 1.000 & 1.000 \\
 & GetWeather          & 1.000 & 1.000 & 1.000 \\
 & PlayMusic           & 0.979 & 0.979 & 0.957 \\
 & RateBook            & 1.000 & 1.000 & 1.000 \\
 & SearchCreativeWork  & 0.905 & 0.952 & 0.929 \\
 & SearchScreeningEvent& 1.000 & 1.000 & 1.000 \\
\midrule

\multirow{7}{*}{Gemini baseline}
 & AddToPlaylist       & 0.99 & 0.96 & 0.99 \\
 & BookRestaurant      & 1.00 & 1.00 & 1.00 \\
 & GetWeather          & 1.00 & 0.99 & 1.00 \\
 & PlayMusic           & 0.97 & 0.97 & 0.95 \\
 & RateBook            & 0.99 & 0.99 & 0.99 \\
 & SearchCreativeWork  & 0.88 & 0.93 & 0.91 \\
 & SearchScreeningEvent& 0.99 & 1.00 & 0.99 \\
\bottomrule
\end{tabular}
\caption{SNIPS accuracy by intent and language condition (descriptive intent-level summaries, not an unweighted decomposition of full-run micro accuracy). EN denotes
English, UR-TR denotes translated Urdu, and UR-CS denotes code-switched
Urdu.}
\label{tab:snips-intent-full}
\end{table*}

\FloatBarrier
\end{document}